\documentclass[10pt,twocolumn,letterpaper]{article}

\usepackage[pagenumbers]{cvpr} 

\definecolor{cvprblue}{rgb}{0.21,0.49,0.74}
\usepackage[pagebackref,breaklinks,colorlinks,allcolors=cvprblue]{hyperref}

\usepackage{CJKutf8} 
\usepackage{url}            
\usepackage{amsfonts}       
\usepackage{amsmath}       
\usepackage{xcolor}         
\usepackage{graphicx}
\usepackage{amssymb, subcaption, overpic, textpos, xfrac}
\usepackage{bm}
\usepackage{multirow}       
\definecolor{lightergray}{gray}{0.9}

\usepackage[utf8]{inputenc} 
\usepackage[T1]{fontenc}    
\usepackage{booktabs}       
\usepackage{nicefrac}       
\usepackage{microtype}      
\usepackage{colortbl}
\usepackage{stfloats}
\usepackage{pifont}
\usepackage{tikz}

\usepackage{adjustbox} 
\usepackage{wrapfig} 

\usepackage{textcomp}  
\usepackage{scalerel}  

\def\drink{\scalerel*{\includegraphics{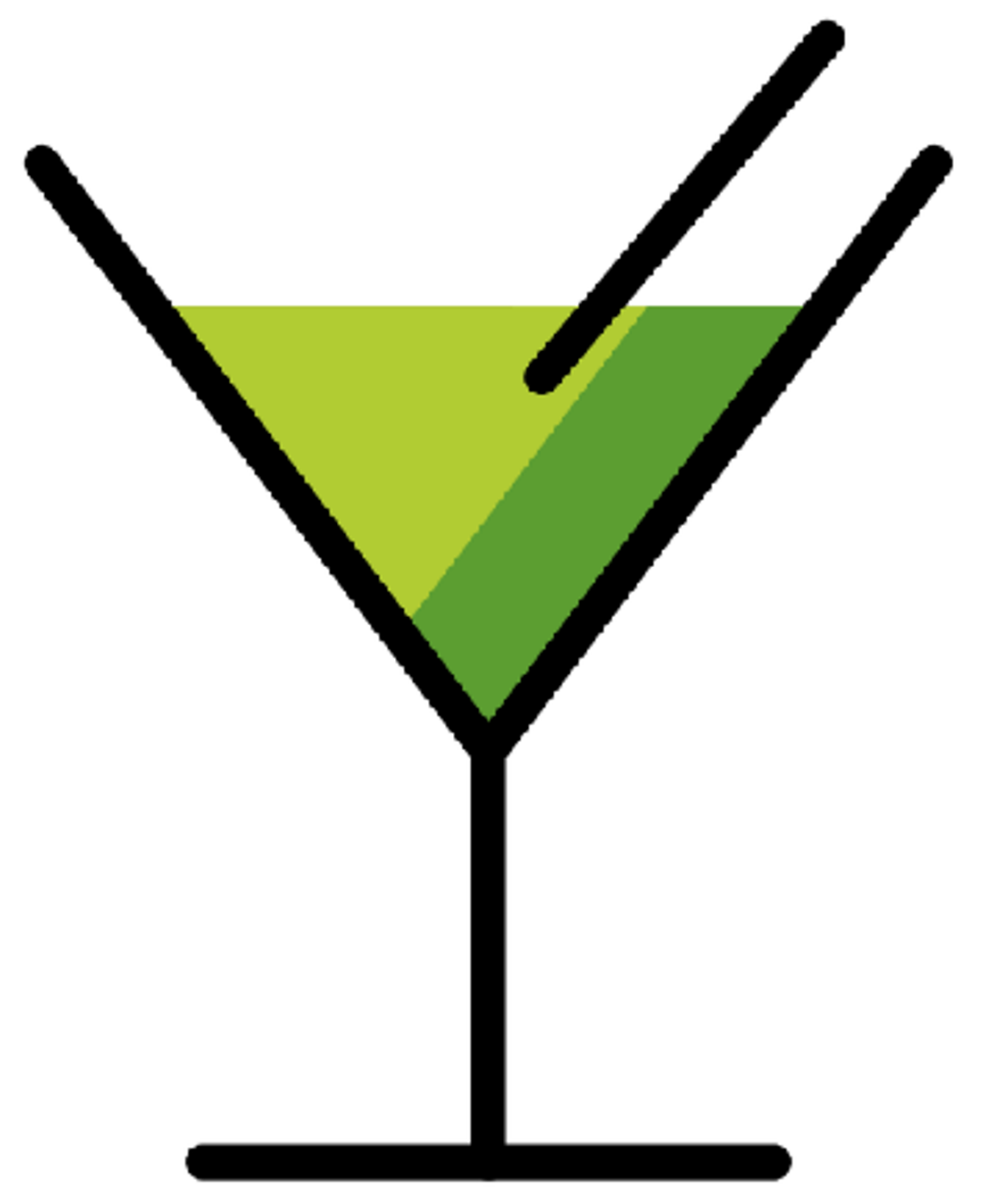}}{\textrm{\textbigcircle}}}
\def\lime{\scalerel*{\includegraphics{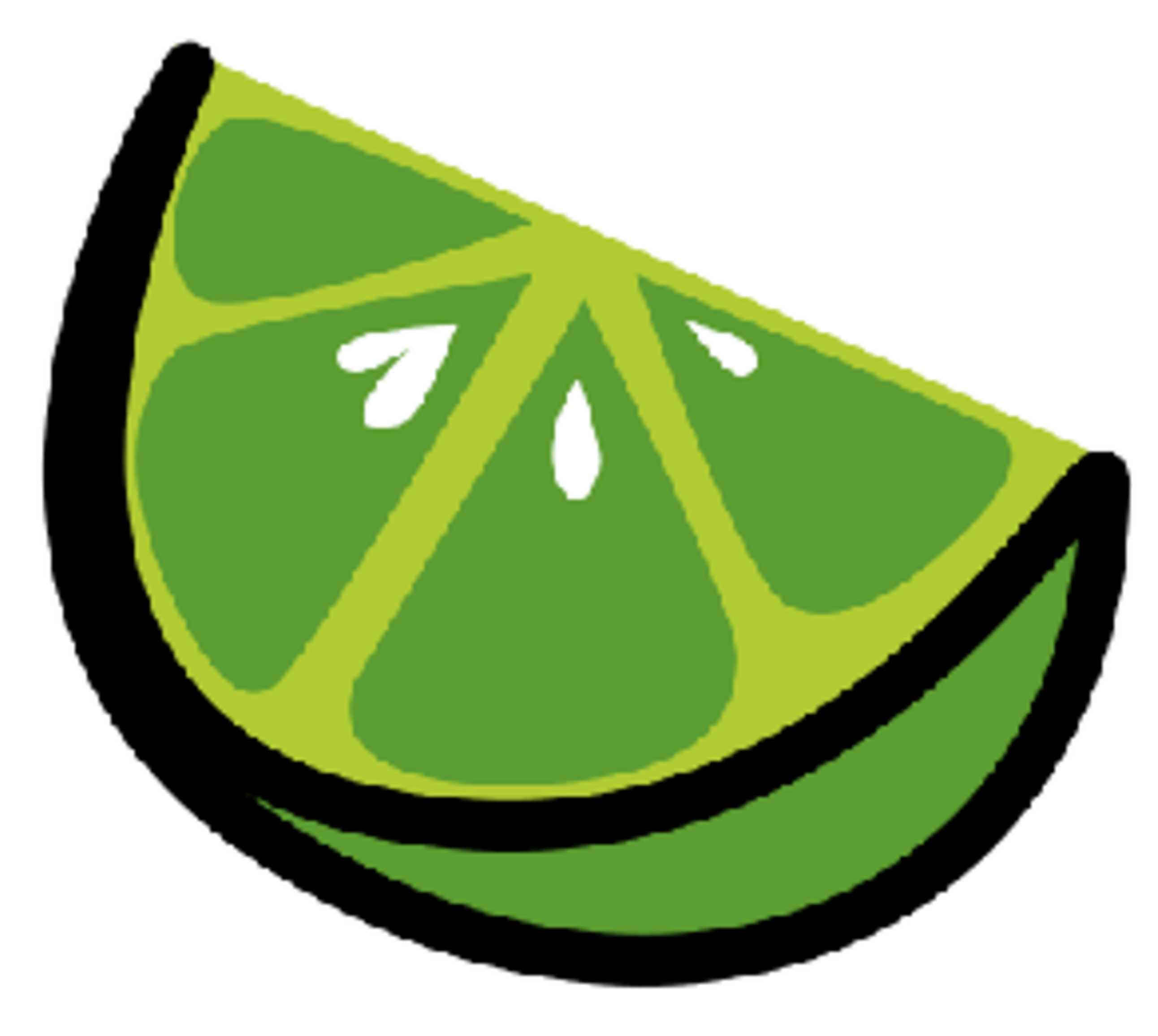}}{\textrm{\textbigcircle}}}

\definecolor{deemph}{gray}{0.6}
\newcommand{\gc}[1]{\textcolor{deemph}{#1}}
\definecolor{baselinecolor}{gray}{.9}

\definecolor{dt}{HTML}{ADCAD8}
\definecolor{dt2}{HTML}{cddfe7}

\newcommand{\otsmodel}[1]{\cellcolor{dt2}{#1}}

\newcommand{\tablestyle}[2]{\setlength{\tabcolsep}{#1}\renewcommand{\arraystretch}{#2}\centering\footnotesize}

\newcolumntype{x}[1]{>{\centering\arraybackslash}p{#1pt}}
\newcolumntype{y}[1]{>{\raggedright\arraybackslash}p{#1pt}}
\newcolumntype{z}[1]{>{\raggedleft\arraybackslash}p{#1pt}}

\definecolor{defaultcolor}{HTML}{E8E2F7}
\newcommand{\default}[1]{\cellcolor{defaultcolor}{#1}}

\newcommand{\dataGovernance}{data governance}
\newcommand{\dataGovernor}{data governor}

\newcommand{\datajuicer}{DataJuicer}

\newcommand{\datasieve}{DataSieve}

\def\paperID{05247} 
\def\confName{ICCV}
\def\confYear{2025}

\title{\lime{} $\rightarrow$ \drink{}: Squeeze Out Tokens from Sample for Finer-Grained Data Governance}

\author{
Weixiong Lin$^{1,2}$, 
Chen Ju$^{1}$\textsuperscript{\ding{41}}, 
Haicheng Wang$^{1,2}$, 
Shengchao Hu$^{2}$,  
Shuai Xiao$^{1}$\textsuperscript{\ding{41}}, \\ 
Mengting Chen$^{1}$, 
Yuheng Jiao$^{1}$, 
Mingshuai Yao$^{1}$,
Jinsong Lan$^{1}$, 
Qingwen Liu$^{1}$,
Ying Chen$^{1}$
\\
$^1$ Alibaba Group \ \ \quad
$^2$ Shanghai Jiao Tong University  \\
{\tt\small wx\_lin@sjtu.edu.cn, cju.void@gmail.com, \{shuai.xsh, yingchen\}@gmail.com}
}

\begin{document}
\maketitle

\begin{abstract}

Widely observed data scaling laws, in which error falls off as a power of the training size, demonstrate the diminishing returns of unselective data expansion.
Hence, data governance is proposed to downsize datasets through pruning non-informative samples.
%
Yet, isolating the impact of a specific sample on overall model performance is challenging, due to the vast computation required for tryout all sample combinations.
Current data governors circumvent this complexity by estimating sample contributions through heuristic-derived scalar scores, thereby discarding low-value ones.
Despite thorough sample sieving, retained samples contain substantial undesired tokens intrinsically, underscoring the potential for further compression and purification.
%
In this work, we upgrade data governance from a `sieving' approach to a `juicing' one.
Instead of scanning for least-flawed samples, our dual-branch \datajuicer{} applies finer-grained intra-sample governance.
%
%
%
%
%
It squeezes out informative tokens and boosts image-text alignments.
Specifically, the vision branch retains salient image patches and extracts relevant object classes,
while the text branch incorporates these classes to enhance captions.
Consequently, \datajuicer{} yields more refined datasets through finer-grained governance.
%
Extensive experiments across datasets demonstrate that \datajuicer{} significantly outperforms existing \datasieve{} in image-text retrieval, classification, and dense visual reasoning.



\end{abstract}

\section{Introduction}

\begin{figure}[t]
\centering
\includegraphics[width=0.47\textwidth]{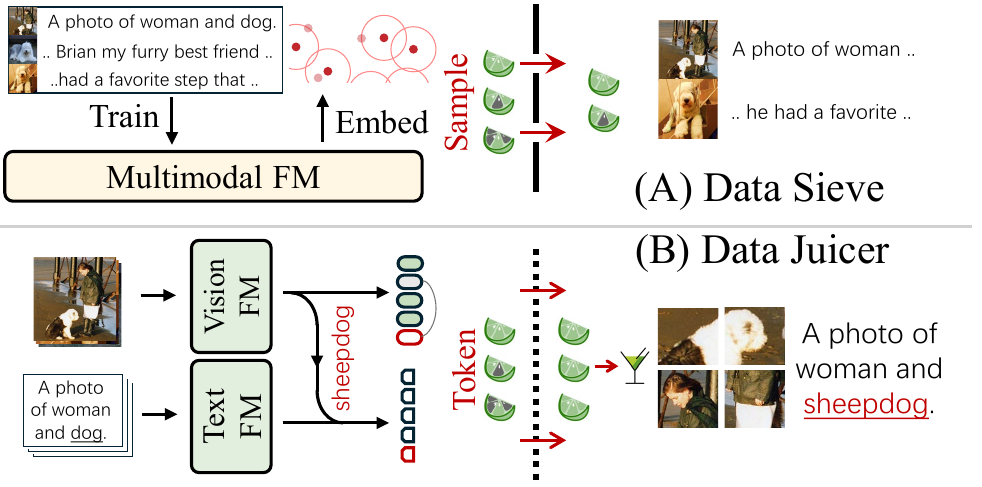}
\vspace{-0.3cm}
\caption{
\textbf{(A) Data Sieve} relies on multi-modal foundation models(FMs) to retain high-value samples.
The hurdle of sample value estimation necessitates the use of manually-designed strategies for governance, limiting Data Sieve's generalizability across datasets.
\textbf{(B) Data Juicer} employs the vision FMs to retain informative image patches, and the text FMs to enhance captions by incorporating visual semantics.
The automatic pipeline yields more accurate and generalizable sample refinement through finer-grained governance. 
}
\label{fig:teaser}
\end{figure}


Vision-language pretraining
(VLP)~\cite{zhang2024vision,jia2021scaling,lin2023pmc,wang2024advancing} is the cornerstone of foundation models (FM)~\cite{gpt-4v,liu2024improved,bai2023qwen} during their pursuing for artificial general intelligence (AGI)~\cite{goertzel2014artificial,baum2017survey}.
CLIP~\cite{radford2021learning} as an example of vision-language models (VLMs), used as default initialization in AIGC models~\cite{ho2020denoising,tian2025visual},
multi-modal large language models (MLLMs)~\cite{zhang2023pmc,chen2024internvl}, etc.
Despite the essential role that large-scale web data now play in VLP,
unselective data expansion hits diminishing returns due to the power-law nature of scaling laws~\cite{sorscher2022beyond}.
Excess samples contribute minimal novel information while potentially introducing noise, leading to poor performance on various downstream tasks.

To address this challenge, \dataGovernance{} is proposed to elitize the dataset by selecting only informative samples for training, thereby optimizing model performance (Tab.~\ref{tabs:data_governance_compare}).
Coreset selection, for instance, develops score functions and discards superfluous data points, though these have largely been at sub-ImageNet scale~\cite{toneva2018empirical,paul2021deep}.
Data pruning aims at large-scale \dataGovernance{} and develops self-supervised pruning metrics, yet primarily evaluated on vision-only models~\cite{sorscher2022beyond}.
%
\textbf{\datasieve{}} extends similar idea to multi-modal data,
prioritize image-text pairs selection according to samples' contributions
~\cite{wang2023too,sorscher2022beyond,tirumala2024d4,abbas2023semdedupdataefficientlearningwebscale}.

Yet, isolating the impact of samples on overall model performance
is challenging.
Current \datasieve{}s circumvent this complexity by using \underline{heuristic-based} metrics as an approximation (\textit{e.g.} the distance to near centroid, similarity with neighbors)~\cite{tirumala2024d4}.
Despite \underline{sample-wise} sieving for de-duplication and diversification,
the substantial undesired image pixels and caption words within retained samples are unaddressed~\cite{wang2023too}.
%
Also, image-text alignment is crucial to VLP,
while \datasieve{} is merely choosing the best from inherently \underline{misaligned} web data.
%
Such coarse-grained \dataGovernor{}s
manage to eliminate easily identified low-quality samples, but their heuristic nature also leads to biases.
While beneficial at smaller scales, the gains of \datasieve{} erode with increasing data scale and may even turn detrimental~\cite{goyal2024scaling}.

Motivated by these observed limitations of \datasieve{},
we take one step back and ask,
is it possible to combat these limitations with finer-grained \dataGovernance{}?

\newlength\savewidth\newcommand\shline{\noalign{\global\savewidth\arrayrulewidth
  \global\arrayrulewidth 1pt}\hline\noalign{\global\arrayrulewidth\savewidth}}

\newcommand*\colorcmark[1]{%
  \expandafter\newcommand\csname #1cmark\endcsname{\textcolor{#1}{\ding{51}}}%
}
\colorcmark{green}

\newcommand*\colorxmark[1]{%
  \expandafter\newcommand\csname #1xmark\endcsname{\textcolor{#1}{\ding{55}}}%
}
\colorxmark{red}

\begin{table*}[t]
\footnotesize
\setlength{\tabcolsep}{7.6pt} 
\centering
\begin{tabular}{@{}lllllccl@{}}
    Method~~~~~~~~~~~~~~~~~~~~~~~~~~~ & Data Type~~~~~~~~~~~~  & Supervision~~~~~~~~~~~~  & Granularity~~~~~~~~  & Estimation  & Alignment & Large-scale & Year \\
    \shline
    Coreset Selection~\cite{paul2021deep} & Image & Class Label  & Sample & heuristic & \textcolor{lightgray}{-} & \redxmark & 2021 \\
    Data Pruning~\cite{sorscher2022beyond} & Image & Self Supervision & Sample & heuristic & \textcolor{lightgray}{-} & \greencmark & 2022 \\
    \datasieve{}~\cite{wang2023too,tirumala2024d4,abbas2023semdedupdataefficientlearningwebscale} & Multi-modality & Image-text Pairs & Sample & heuristic & \redxmark & \greencmark & 2024  \\
    \datajuicer{} (\textbf{Ours}) & Multi-modality & Image-text Pairs & Patch / Token & model-derived & \greencmark & \greencmark & -  \\
\end{tabular}
\vspace{-1em}
\caption{\label{tabs:data_governance_compare} 
\textbf{Data Governance Methods}.
Our \datajuicer{} differs from \datasieve{} in granularity.
It extends \dataGovernance{} to image-patch/word-token level.
With model-derived contribution estimation,
\datajuicer{} can smoothly scale to large data spared from heuristic-induced biases.
We further improve image-text alignment by enhancing captions with visual evidence.
}
\vspace{-1em}
\end{table*}

Our exploration upgrades data governance from a sieving approach to a `juicing’ one (Fig.~\ref{fig:teaser}).
Instead of scanning for least-flawed samples, 
our \datajuicer{} cracks each sample (image, caption) into token-level ingredients (patches, words).
%
%
We employ dual-branch to squeeze out informative tokens only.
(1) \textit{Vision Branch} uses Vision FMs ({\em e.g.} DINO~\cite{caron2021emerging}) to patchify the image and perform \underline{token-wise} reduction.
Most contributive patches are retained based on \underline{model-derived} estimation from Vision FMs' attention mechanism.
Vision FMs also identify relevant object classes present in the image.
(2) \textit{Text Branch} use Language FMs ({\em e.g.} LLaMA) to correct text imperfections such as misspellings and repetitive expressions.
Also, the visual evidences from the vision branch are incorporated,
we use high-confidence object classes to rectify captions for \underline{better-aligned} image-text pairs.
Note that \datasieve{} and \datajuicer{} are complementary rather than mutually exclusive, and their combined use can further enhance data governance.

We show through comprehensive experiments that \datajuicer{} significantly enhances data efficiency during training and improves performance on over ten downstream tasks.
Furthermore, our findings indicate that \datajuicer{} performs effectively without incurring computational overhead.
Thorough ablations are done to reveal each component’s effectiveness, both quantitatively and qualitatively.

To sum up, our contribution lies in three folds:
\begin{enumerate}
    \item We propose juicing-based \dataGovernor{}s
    different from sieving-based ones,
    providing a finer-grained pipeline to produce high-quality image-text pairs.
    \item We design \datajuicer{} to reduce undesired tokens and enhance cross-modal alignment, by using vision/language FMs for automatic model-derived token-value estimation.
    \item We conduct extensive experiments and ablations on many benchmark datasets, demonstrating the data improvements achieved by our \datajuicer{} and showcasing the superior performance of trained VLMs.
\end{enumerate}

%
%
\section{Related Works}
\label{sec:related work}
\noindent \textbf{Foundation Models (FMs)} trained on large-scale datasets have significantly advanced progress in both CV and NLP. 
FMs are generally classified into two categories: uni-modal FMs and multi-modal FMs.
(1) Uni-modal FMs capture deep, domain-specific knowledge by training on large-scale intra-modal data, such as DINO \cite{caron2021emerging} and LLaMA \cite{touvron2023llama}.
(2) Multi-modal FMs have seen rapid advancements, with notable models including GPT-4V \cite{gpt-4v} and LLaVA \cite{liu2024improved}.
Uni-modal FMs play a crucial role in multi-modal FMs by providing specialized, in-depth knowledge within their domains.

\begin{figure*}[t]
\centering
\includegraphics[width=1.0\textwidth]{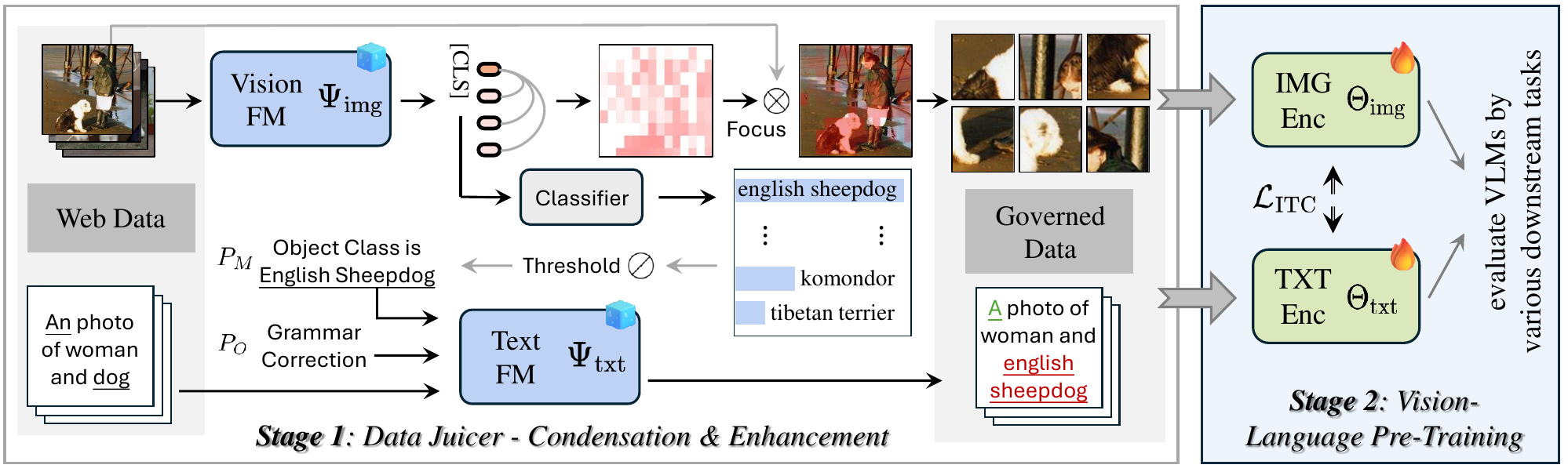}
\vspace{-0.5cm}
\caption{\textbf{Pipeline Overview.} Our framework employs vision and text branches to reduce undesired ingredients in real-world data. 
Vision Foundation Model (FM) is used to discard image patches with low contribution to overall semantics and to extract object classes.
Text Foundation Model (FM) rectify grammatical errors and refine the textual descriptions
By incorporating high-confidence class names, the text branch enhances image-text consistency.
This process results in data that is both less redundant and less noisy, therefore leading to more effective training of Vision-Language Pretraining.}
\label{fig:framework}
\end{figure*}

\vspace{0.1cm}
\noindent \textbf{Vision-Language Pretraining (VLP)} establish cross-modal alignment by leveraging large-scale web-sourced datasets, as exemplified by models such as CLIP~\cite{radford2021learning}, ALIGN~\cite{jia2021scaling}, Florence~\cite{yuan2021florence}, FILIP~\cite{yao2021filip}, and PMC-CLIP~\cite{lin2023pmc}.
VLP architectures are grouped into three types: single tower~\cite{chen2020uniter,kim2021vilt}, twin towers~\cite{radford2021learning,jia2021scaling}, and bridge tower~\cite{li2023blip,li2022blip,zhu2023minigpt}.
In terms of optimization techniques, contrastive learning~\cite{chen2020learning,zheng2021contrastive} and cross-modal matching~\cite{cheng2023vindlu,li2022blip} dominate, with applications ranging from self supervision~\cite{liu2022exploiting,liu2024annotation}, weak supervision~\cite{li2023blip,cheng2023mixer,cheng2023category}, to partial supervision~\cite{ju2023constraint,liu2024audio}. 
As a foundation for multi-modal FMs, VLPs contribute to various downstream tasks, including image perception~\cite{feng2022promptdet,minderer2022simple,ju2023turbo,ju2025turbo,cheng2023image,chen2023enhancing}, video understanding~\cite{ju2021divide,liu2024audio,cheng2024denoiser,zhao2020bottom,ju2023distilling,ju2022adaptive,ju2020point,liu2023audio,wang2025contrast}, open-vocabulary learning~\cite{ju2022prompting,yang2024multi,ju2023multi,ma2024open,ma2023open,yang2023multi}
and AIGC generation~\cite{ma2023diffusionseg,chen2024wear,ma2025freesegdiff,wang2025folder}. 
Notably, most multi-modal FMs initialize their visual encoders using pre-training like CLIP.

\vspace{0.1cm}
\noindent \textbf{Data Governance}.
Large-scale image-text data are crucial for building powerful VLMs, which serve as the foundation for multi-modal FMs.
While VLPs consume data at a rapid rate, far exceeding the volume that can be manually curated, necessitating the use of automated data pipelines for web-collected data~\cite{schuhmann2022laionb}.
However, web-sourced data often suffer from noise and redundancy, leading to suboptimal VLP performance.
As in Tab.~\ref{tabs:data_governance_compare},
recent \dataGovernance{} explores to address these challenges
~\cite{wang2023too,tirumala2024d4,abbas2023semdedupdataefficientlearningwebscale}.
These approaches break through the vision-only constraints of Coreset Selection and Data Pruning by
sieving out contributive image-text pairs with heuristic-estimated informativity,
a framework we term \datasieve{}.
De-duplication and diversification are two main goals of \datasieve{}.
It employs the visual embeddings of samples for clustering
and discards those proximal to centroids to reduce semantic overlapping samples similarity~\cite{abbas2023semdedupdataefficientlearningwebscale}.
Dedicate sampling strategies are also developed to mine neighboring relationships among samples, thereby enhancing data diversity~\cite{maharana2023d2}.
DataSieves focus on scanning least-flawed samples while ignoring the image-text misalignment.
Some attempts to synthesize captions with MLLM~\cite{wang2023too,zhang2025long,fan2024improving,lai2025veclip},
though effective, the generated captions tend to be homogenized, reducing the diversity and richness of the provided information when more samples are modified.
The use of MLLMs for captioning actually introduces a paradox of mutual dependence between VLMs and MLLMs:
Advancements in VLMs are prerequisites to enhancing the recaptioning capabilities of MLLMs.
Furthermore, these models tend to prioritize text, often overlooking redundancy within image pixels.

\section{Methods}

\subsection{Notations and Preliminaries}

\newcommand{\datasetTXT}{D_\text{txt}}
\newcommand{\datasetIMG}{D_\text{img}}
\newcommand{\Img}{I}
\newcommand{\Txt}{T}
\newcommand{\governor}{\Psi}
\newcommand{\governorTXT}{\Psi_\text{txt}}
\newcommand{\governorIMG}{\Psi_\text{img}}
\newcommand{\model}{\Theta}
\newcommand{\modelIMG}{\Theta_\text{img}}
\newcommand{\modelTXT}{\Theta_\text{txt}}
\newcommand{\cost}{C}
\newcommand{\costGovern}{C_\text{gv}}
\newcommand{\costTrain}{C_\text{tr}}


\textbf{Task Formulation}.
Given training budgets $\mathcal{T}$,
\dataGovernor{} $\Psi$ is to construct a governed dataset $\mathbb{V}$ from a source dataset with overhead $\mathcal{S}$, where $\mathbb{D} = \{I_i,T_i\}_{i=1}^N$ comprises $N$ image-caption pairs.
And then optimize model $\Theta$ on vision-language pre-training task $\mathcal{L}$,
thereby maximizing general performance $\mathcal{A}$.
While $\mathcal{A}$ is challenging to verify on a given vision-language model,
it can be inferred through diverse abilities such as visual recognition and compositional reasoning, with evaluation tools like VLMEvalKit~\cite{duan2024vlmevalkit}.
Mathematically $\mathbb{V} = \Psi ( \mathbb{D} )$, and $\Psi$ can be obtained as
\begin{equation}
    \arg\max_\Psi \mathcal{A}(\arg\min_{\Theta} \mathcal{L}(\Theta, \mathcal{T}, \Psi)),
\end{equation}
%
%
There are various options to define $\mathcal{T,S,A}$ and $\mathcal{L}$.
In this work, we define the training budget $\mathcal{T}$ as the number of pre-trained tokens, and overhead $\mathcal{S}$ as the time investment in data construction.
The pre-training objective $\mathcal{L}$ is decided by VLP (InfoNCE, Sigmoid, Caption Loss, etc~\cite{radford2021learning,zhai2023sigmoid,yu2022coca}).
And the vision-language model's performance $\mathcal{A}$ as the evaluation across different tasks using the VLMEvalKit framework.

%
\vspace{0.1cm}
\noindent\textbf{Motivation}.
Although images $I$ and texts $T$ appear symmetrical, they differ in underlying semantic structures.
Images consist of abundant pixels that convey precise semantics, where many patches contribute little to the overall meaning and can be discarded.
In contrast, text is organized concisely, but its semantics are often vague and prone to disruption by noise.
%
While \datasieve{} overlooks
the semantic discrepancy which hinders vision-language pretraining.
This observation motivates us to design a finer-grained \dataGovernor{} which can effectively balance both modalities, {\em i.e.}, reduce image redundancy and enhance text semantics, thus improving data quality and facilitating more efficient training.


\vspace{0.1cm}
\noindent\textbf{Pipeline Overview}. 
Our \dataGovernance{} consists of two parallel branches in Fig.~\ref{fig:framework}, named \textbf{vision branch} and \textbf{text branch}. 
The vision branch reduces redundancy by discarding semantically insignificant image patches and assesses the probability of object presence within the scene. This process effectively discretizes image semantics.
The text branch then leverages visual objects as guidance for caption enhancement, achieving greater clarity and precision.
Through these two branches, we generate a modified dataset $\mathbb{V}$ comprising image-text pairs that are less redundant and better aligned.
Pretrained models $\Theta$ are subsequently applied to various downstream tasks, whose performances serve as a measure of the effectiveness of the \dataGovernor.

\subsection{Data Juicer: Condensation \& Enhancement}

\newcommand{\patchIT}{\Img_i^t}
\newcommand{\uniGovIMG}{\Gamma_\text{img}}
\newcommand{\uniGovPred}{\Gamma_\text{pred}}
\newcommand{\uniGovTXT}{\Gamma_\text{txt}}

We aim to reduce image redundancy at the patch level and improve cross-modal consistency.
%
Given a dataset $\mathbb{D} = (\Img_i, \Txt_i)_{i=1}^N$, where $(\Img, \Txt)$ represents a raw image-text pair, we first patch the image $\Img$ to remove redundant patches and then enhance the text $\Txt$ with visual semantics from $\Img$. 
The vision and text branches are outlined as follows.
%
%

\vspace{0.1cm}
\noindent\textbf{The Vision Branch} reduces image patch redundancy by evaluating each patch's contribution to the overall semantics, using similarity scores to identify and discard insignificant patches. 
Specifically, the vision governor $\governorIMG$ eliminates patches whose embeddings significantly diverge from the [CLS] embedding, which encapsulates the image's global semantics.
By removing patches with low similarity to the [CLS] embedding, redundancy is effectively reduced. 
Additionally, the class prediction derived from the [CLS] embedding provides object existence probabilities, guiding the enhancement of textual descriptions.

With image $\Img\in \mathbb{R}^{H\times W}$ split into $m$ patches
$\{\Img_t\}_{t=1}^m \in \mathbb{R}^{m\times B}$, each of $B = HW / m^2$ pixels.
$\governorIMG$ scores and selects patches with its last-layer attention mechanism.
We initialize $\governorIMG$ with DINO~\cite{caron2021emerging},
and denote $\bm{W}_Q, \bm{W}_K \in \mathbb{R}^{d_k}$ as query, key matrix.
From previous layer, there are corresponding visual tokens $\{\bm{v}_t\}_{t=1}^m \in \mathbb{R}^{m\times d}$, and a special [CLS] token $\bm{v}_c \in \mathbb{R}^d$ representing the overall feature of $\Img$.
The contribution scores $\bm{s}\in \mathbb{R}^m$ of the image patches ${\Img_t}$  are determined by 
\begin{equation}
 \bm{s} = \frac{
 (\bm{v}_c \bm{W}_Q)
 ([\bm{v}_1,\dots,\bm{v}_m] \bm{W}_K)^T
 }{\sqrt{d_k}}
\end{equation}
Thus the top-k patches are retained because of the most contribution to the global visual embedding $\bm{v}_c$.
While the others, contributing less to the overall semantics, are discarded.
This reduction in visual tokens reduces the computational cost during VLP.
The resulting image is represented as:
\begin{equation}
\Img' = \{\Img_t\}_{t=1}^k \in \mathbb{R}^{k\times B}.
\end{equation}

Given the image embedding $\bm{v}_c$, the vision governor $\governorIMG$ predicts $L$ class labels $C$ of the image $\Img$ using a class prediction head, $\uniGovPred$ initialized from DINO.
The object existence probabilities are represented as $\bm{p} \in \mathbb{R}^L$.
\begin{equation}
\bm{p} = \uniGovPred(\bm{v}_c).
\label{equation: class_pred}
\end{equation}

\noindent\underline{\textit{Remark}.} In the vision branch, Vision FMs can offer us free knowledge about data requiring no extra computation cost.
The vision governor reduces redundancy by discarding noisy, insignificant patches while retaining the most informative ones. As a result, the new image $\Img'$ is of higher quality and less computation for $\Theta$ pretraining.
Furthermore, the class prediction process
transforms redundant visual semantics into discrete object classes, offering clear visual evidence for the text branch's further enhancement.

\newcommand{\promptObj}{P_O}
\newcommand{\promptMistake}{P_M}

\vspace{0.1cm}
\noindent\textbf{The Text Branch}.
To mitigate the noise impact in input texts, the text governor $\governorTXT$ enhances and denoises texts $\Txt$ with the assistance of visual semantics.
$\governorTXT$ is one Large Language Model, such as LLaMA~\cite{touvron2023llama}, Qwen~\cite{bai2023qwen},
and we construct the prompts $\promptObj$, incorporating class names given by $\bm{p}$ from Equ.~\ref{equation: class_pred}.
As the distribution of $\bm{p}$ indicates the confidence of class predictions, we set a threshold $\epsilon$ and 
only take classes of $\bm{p} > \epsilon$ into $P_O$ construction, denoted as $C'$.

As texts $\Txt$ may contain imperfections like misspellings and repetitive expressions, the text governor $\governorTXT$ leverages contextual information to denoise the text. 
It constructs a textual correction prompt $\promptMistake$
and combine it with the visual correction prompt $\promptObj$.
As a result, $\governorTXT$ uses both $\promptMistake$ and $\promptObj$ as prompts,
to correct the imperfections in $\Txt$,
thereby generating new caption $\Txt'$.

\begin{equation}
    T'_i = \governorTXT(\promptMistake, \promptObj, T_i).
\end{equation}
Combining the efforts of both vision and text branches, we get new training data
\begin{equation}
    \mathbb{V} = \{\Img_i', \Txt_i'\}_{i=1}^N.
\end{equation}

\noindent\underline{\textit{Remark}.}
In the text branch, we use the LLMs to enhance and denoise the text data:
(1) \textit{Enhancement}:
Incorporating visual evidences $C'$ from the vision branch $\governorIMG$ is challenging,
due to the absence of direct mapping from $C'$ to raw caption components in $\Txt$.
%
%
%
LLMs $\governorTXT$, leveraging inherent world knowledge,
infer the linking from visual evidence to caption components, thereby enhancing the captions.
(2) \textit{Denoising}: Web-sourced texts are often informal and grammatically lax,
rendering them unsuitable for image descriptions.
The text governor $\governorTXT$ performs textual correction to eliminate errors such as misspellings and repetitive expressions
to improve the captions' quality.

To conclude, with the vision and text branch combined, we obtain a more compact dataset, which has less noise and improved image-caption consistency.

\subsection{Vision-Language Pre-Training}
\newcommand{\imgEmb}{\bm{v}_\text{img}}
\newcommand{\txtEmb}{\bm{v}_\text{txt}}
Given the governed dataset $\mathbb{V}$, we train vision-language models $\Theta$ from scratch. We describe the architecture first and then introduce the training objectives.

\vspace{0.1cm}
\noindent \textbf{Architecture.} Given $\mathbb{V} = (\Img_i', \Txt_i')_{i=1}^N$ as image-text training pairs,
the vision-language pretraining models $\Theta$ typically comprises
an image encoder $\modelIMG$ and a text encoder $\modelTXT$.
We use CLIP~\cite{radford2021learning} for demonstration.

In detail,
we encode a specific image-text pair $(\Img, \Txt)$ separately with the image/text encoder $\model_\text{img/txt}$, the embedding dimension is denoted as $d$:
\begin{align}
   \imgEmb' &= \modelIMG(\Img') \in \mathbb{R}^d, \\
   \txtEmb' &= \modelTXT(\Txt') \in \mathbb{R}^d,
\end{align}
where $\imgEmb'$ represents the embedding for the image, and $\txtEmb'$ refers to the text embedding.

\vspace{0.1cm}
\noindent \textbf{Image-Text Contrastive Learning~(ITC).}
We implement ITC loss following CLIP~\cite{radford2021learning}, which aligns the corresponding visual and textual representations for each sample in a batch.
In detail, denoting batch size as $b$, we calculate the softmax-normalized cross-modality dot product similarity between the current visual/text embedding~($\imgEmb'$/$\txtEmb'$) and all samples within the batch, termed as $p^{\text{i2t}}, p^{\text{t2i}} \in \mathbb{R}^{b}$, and the final ITC loss is formulated as:
%
%
\begin{equation}
    \mathcal{L}_{\text{ITC}} = \mathbb{E}_{(I, T)\sim \mathbb{V}} \big[ \text{CE}( y^{\text{i2t}}, p^{\text{i2t}} ) + \text{CE}( y^{\text{t2i}}, p^{\text{t2i}} ) \big],
\end{equation}
where $ y^{\text{i2t}}, y^{\text{t2i}}$ refer to one-hot matching labels, and $\text{CE}$ refers to the InfoNCE loss~\cite{he2020momentum}.

\vspace{0.1cm}
\noindent \textbf{Train \& Inference.}
While image patches used during training and inference differ, our framework is flexible and requires no modifications for downstream tasks. The vision-language model is pre-trained on a subset of image patches, yet it can be directly applied to complete images during inference. 
This straightforward setup yields competitive performance on downstream tasks and serves as a baseline for our ablation experiments.

Using a subset of image patches as input during inference is also allowed.
The patch reduction creates a trade-off between inference speed and performance.
For downstream tasks requiring only coarse-grained visual semantics, they can use fewer patches as input for inference acceleration.


\section{Experiment}

\subsection{Datasets \& Implementation Details}

\noindent \textbf{Training Datasets}.
\underline{CC3M}~\cite{sharma2018conceptual} and \underline{CC12M}~\cite{changpinyo2021cc12m} are widely used vision-language pre-training datasets.
They gather massive web-sourced images and alt-texts, then perform simple data cleansing, {\em e.g.}, image filtering by resolution, and text cleaning by length.
Despite their vast scale and rich diversity, unhelpful or even detrimental samples persist.
\underline{YFCC15M}~\cite{radford2021learning}, a 15M subset of the multilingual and noisy YFCC100M~\cite{thomee2016yfcc100m} that contains English captions, is also used to validate \datajuicer's generalizability to data in the wild. LAION400M~\cite{schuhmann2022laionb} is a even larger noisy dataset for VLP.
To reduce the computation cost and storage overhead, we randomly sample \underline{LAION40M} from LAION400M.

\vspace{0.1cm}
\noindent \textbf{Testing Datasets \& Metrics}.
For image-text retrieval, MSCOCO~\cite{lin2014microsoft} and Flickr30~\cite{young2014image}  are used as standard benchmarks.
For image classification, we utilize the ImageNet~\cite{deng2009imagenet}(ImgN1K/-R/-A) validation set.
To demonstrate how \datajuicer{}-enhanced data improves visual encoder,
We replace LLaVA-13B~\cite{liu2024improved}'s vision backbone with our image encoder pretrained on refined datasets $\mathbb{V}$ for comprehensive evaluation.
We follow conventions to evaluate with consensus metrics: using Top-1/5 accuracy for classification; R@1/5/10 for image-text retrieval; and follow~\cite{duan2024vlmevalkit} to evaluate LLaVA on~\cite{liu2025mmbench,li2024seed,lu2022learn,schwenk2022okvqa,singh2019towards,Guan_2024_CVPR,yu2024mm,fu2023mme,mishraICDAR19}.
Note that all testing are standardized to ensure fairness.

\vspace{0.1cm}
\noindent\textbf{Implementation}
We utilize PyTorch~\cite{paszke2019pytorch} to implement our models and train them on 8 NVIDIA A100 GPUs.
The model is pre-trained for 50 epochs with a batch size of 1024 and an AdamW~\cite{loshchilov2017decoupled} optimizer with a weight decay of 0.05.
During training, we apply a learning rate warm-up to 3e-4 and OneCycle scheduler (cosine annealing)~\cite{loshchilov2016sgdr}.
All mages were cropped to a resolution of 224$\times$224 during training and inference.
More training hyperparameters for downstream tasks are put in the supplementary materials.

\subsection{Comparison with SoTA Methods}

\noindent\textbf{Performance Comparisons.}
Tab.~\ref{tab:sota} compares our \datajuicer{} with coarse and fine-grained data governing methods on image-text retrieval (COCO, Flickr), image classification (ImgN1K, ImgNR, ImgNA) and MLLM downstream tasks (OKVQA, CCBench, etc).
In the first part, we introduce \datasieve{}, whose simplest baseline is random selection before training.
Clip-score, as a classical metric for image-text correspondence estimation, is used to decide whether to retain samples.
An alternative baseline clusters the dataset into subsets and enforces uniform sampling across all groups.
To make a wider comparison, we include 4 SoTA sample selection methods~\cite{wang2023too,maharana2023d2,tirumala2024d4,abbas2023semdedupdataefficientlearningwebscale}.
These methods select the most representative samples leveraging embeddings from pre-trained models, via heuristic sampling procedures.
Then we introduce fine-grained governing methods in the second part.
Rather than sample-wise selection based on heuristic metrics, fine-grained \dataGovernor{}s
refine dataset at the token level.
Thus the token-wise Random*/Cluster* approaches produce training samples with higher information density compared to those generated by coarse-grained Random/Cluster operations.
Therefore, as shown in Tab.~\ref{tab:sota}, Random*/Cluster* usually performs better than Random/Cluster and many coarse-grained methods.

From Tab.~\ref{tab:sota}, we obtain two observations:
1). \datajuicer{} shows a clear gain compared to competitors on most downstream tasks; 
2). Fine-grained methods outperform coarse-grained ones even without well-designed sampling.


\newcommand{\blue}[1]{$_{\color{RoyalBlue}\downarrow #1}$}
\newcolumntype{*}{>{\global\let\currentrowstyle\relax}}
\newcolumntype{^}{>{\currentrowstyle}}
\newcommand{\rowstyle}[1]{\gdef\currentrowstyle{#1}#1\ignorespaces}
\definecolor{dt}{gray}{0.7}  %
\newcommand{\gcline}{\rowstyle{\color{dt}}}

\definecolor{lightgreen}{HTML}{D8ECD1}
\newcommand{\better}[1]{\cellcolor{lightgreen}{#1}}

\begin{table*}[t]
    \caption{\textbf{Comparison to State-of-The-Art Methods.}
    All methods set compression ratios to 50\% and train CLIP from scratch for fairness.
    We describe sample-sieving methods as coarse-grained and token-juicing ones as fine-grained.
    \gc{Full Data} refers to results without data governance. More implementation details are available in Appendix~\ref{sec:experiment_details}.
    }
    \label{tab:sota}
    \centering
    \footnotesize
    \setlength{\tabcolsep}{4.6pt} 
    \begin{tabular}{ll|cccccccccccc}
        ~ & Method & COCO & Flickr & ImgN1K & ImgNR & ImgNA & OKVQA & CCBench & HaBench & MMBench & MMMU & MMStar & POPE \\
        \shline
        ~ & \gcline Full Data & \gcline 11.68 & \gcline 26.2 & \gcline 17.71 & \gcline 19.56 & \gcline 4.32 & \gcline 62.00 & \gcline 1.18 & \gcline 35.64 & \gcline 42.35 & \gcline 37.33 & \gcline 26.00 & \gcline 77.84 \\
        \parbox[t]{4mm}{\multirow{7}{*}{\rotatebox[origin=c]{90}{Coarse}}} & Random & 6.18 & 14.7 & 11.28 & 11.99 & 2.52 & 57.11 & 2.33 & 35.43 & 40.63 & 29.30 & 23.20 & 76.02 \\
        ~ & Cluster & 6.58 & 14.6 & 11.28 & 12.08 & 2.56 & 59.56 & 2.15 & 34.06 & 38.83 & 29.30 & 22.66 & \textbf{80.17} \\
        ~ & CLIP-Score & 6.60 & 14.2 & 11.57 & 13.53 & 2.89 & 62.53 & 1.17 & 36.27 & 38.23 & 28.60 & 24.20 & 78.31 \\
        ~ & D2Pruning~\cite{maharana2023d2} & 7.43 & 16.4 & 11.58 & 12.22 & 3.65 & 54.42 & 1.40 & 31.00 & 37.88 & 28.00 & 23.35 & 79.43 \\
        ~ & TLDR~\cite{wang2023too} & 9.66 & 21.4 & 11.88 & 14.79 & 3.28 & 57.9 & 1.18 & 31.96 & 37.20 & \textbf{29.33} & 22.80 & 79.78 \\
        ~ & SemDeDup~\cite{abbas2023semdedupdataefficientlearningwebscale} & 7.06 & 16.4 & 11.55 & 12.64 & 3.40 & 54.23 & 1.37 & 31.54 & 37.28 & 28.00 & 23.46 & 79.51  \\
        ~ & D4~\cite{tirumala2024d4} & 8.83 & 17.9 & 12.25 & 14.12 & 3.93 & 55.88 & 2.69 & 31.59 & 37.93 & 28.30 & 23.30 & 77.16 \\
        \hline
        \parbox[t]{4mm}{\multirow{3}{*}{\rotatebox[origin=c]{90}{Fine}}} & Random* & 9.14 & 21.2 & 14.02 & 15.94 & 3.61 & 57.55 & 2.35 & 36.06 & 38.91 & 29.00 & 24.13 & 77.53  \\
        ~ & Cluster* & 9.42 & 21.4 & 14.86 & 16.08 & 3.56 & 57.86 & 2.43 & 36.21 & 38.67 & 28.70 & 24.50 & 77.29 \\
        ~ & \textbf{\datajuicer} & \textbf{9.72} & \textbf{21.8} & \textbf{15.54} & \textbf{17.32} & \textbf{4.00} & \textbf{62.10} & \textbf{3.52} & \textbf{36.91} & \textbf{42.78} & \textbf{29.33} & \textbf{27.07} & 77.26 \\
    \end{tabular}
\end{table*}

\begin{table*}[t]
\parbox{.52\linewidth}{
    \centering
    \footnotesize
    \caption{
    \textbf{Training Efficiency Comparison.}
    Results are reported on ImageNet-1K with CLIP under 50\% prune ratio on an 8-A100 GPU server.
    \datajuicer{}$\times$Sieve applies \datajuicer{} on retained samples from \datasieve{}.
    Total computation costs are calculated by GPU hours (n*h).
    Our default settings are marked in {\setlength{\fboxsep}{2pt}\colorbox{dt2}{{blue}}.}
    }
    \vspace{-8pt}
    \begin{adjustbox}{max width=\linewidth}

    \begin{tabular}{y{54}x{18}x{20}x{30}x{40}}
        Method & Acc & Time & Overhead & Total {\small (n*h)} \\
        \shline
        \gc{Full Data} & \gc{17.7} & \gc{27} & \gc{-} & \gc{216.1} \\
        TLDR & 11.9 & 14 & 1.2 & 113.2 \\
        D4 & 12.6 & 14 & 1.2 & 113.2 \\
        SemDeDup & 11.6 & 14 & 1.2 & 113.2 \\
        \hline
        \otsmodel{\bf\datajuicer} & \otsmodel{15.5} & \otsmodel{\textbf{13}} & \otsmodel{\bf{0.8}} & \otsmodel{\textbf{104.8}} \\
        \textbf{\datajuicer}$\times$Sieve & \textbf{15.7} & 14 & 2.5 & 114.5 \\
    \end{tabular}
    \end{adjustbox}
    \label{tab:train_efficiency_comparison}
}
\hfill
\parbox{.45\linewidth}{
    \centering
    \footnotesize
    \caption{\textbf{Inference Efficiency Comparison.}
    With image retention ratio $r=\frac{k}{m}=0.5$, our \datajuicer{} exhibits faster inference on equal-sized samples compared to \datasieve.
    We report Top-1 accuracy (acc) on ImageNet-1K and fp32 CLIP(ViT-B/16, BERT-Base) throughput (im/s) on a V100 GPU.
    }
    \vspace{-8pt}

    \begin{adjustbox}{}
    \begin{tabular}{y{52}x{24}x{24}x{24}x{24}}
        Method & Acc & GFlops & Im/s & Speed \\
        \shline
        TLDR & 11.9 & 13.2 & 158 & 1$\times$ \\
        D4 & 12.3 & 13.2 & 158 & 1$\times$ \\
        SemDeDup & 11.6 & 13.2 & 158 & 1$\times$ \\
        \hline
        \otsmodel{\bf \datajuicer$_{r=1.0}$} & \otsmodel{\bf 15.5} & \otsmodel{13.2} & \otsmodel{158} & \otsmodel{1$\times$} \\
        \textbf{\datajuicer}$_{r=.50}$ & 15.4 & 7.7 & 191 & 1.7$\times$ \\
        \textbf{\datajuicer}$_{r=.25}$ & 13.4 & \textbf{4.8} & \bf{585} & \bf 3.7$\times$ \\
    \end{tabular}
    \end{adjustbox}
    \label{tab:infer_efficiency_comparison}
}

\end{table*}

\vspace{0.1cm}
\noindent\textbf{Efficiency Comparisons.}
We also compare efficiency between \datajuicer{} and other methods. 
Although the motivations of these methods are diverse, their main goal is to save training costs.
Thus, we report training time of $\Theta$, overhead used for $\mathbb{V}$ dataset curation, and total GPU hours of these methods in Tab.~\ref{tab:train_efficiency_comparison}.
Previous sieving-based methods discard unworthy samples while ignoring token-wise redundancy.
\datajuicer{} wisely spends computation on most informative tokens told by vision foundation models, further reducing token redundancy.
More importantly, \datajuicer{} achieves both faster inference and superior performance on ImageNet-1K at the image retention ratio $r= k / m = 0.5$.
To explore the joint impact of \datajuicer{} and \datasieve{},
\datajuicer$\times$Sieve applies \datajuicer{} on \datasieve's retained samples. Such one combination further enhances performance and efficiency, thereby confirming \datajuicer’s complementary role in \dataGovernance{} alongside \datasieve.

We show the inference efficiency of \datajuicer{} in Tab.~\ref{tab:infer_efficiency_comparison}.
Finer-grained token view reduces computation cost per sample,
therefore \datajuicer{} exhibits faster speed on equal-sized samples compared to \datasieve{}.
Reducing the retention ratio $r = k / m$
further accelerates inference while incurring a performance trade-off.
This paper sets $r=1.0$ by default, while users may adjust it by need.

\subsection{Scaling Behavior}

To match the extensive training scale of foundation models, we compare scaling behavior between \datajuicer{} and \datasieve.
We demonstrate \datajuicer{}'s superior potential along either of these three axes:

\begin{itemize}
\setlength\itemsep{.1em}
	\item \textbf{Data Scaling}. We scale pre-training data from 3 to 12 million, using the CC12M set \cite{changpinyo2021cc12m}.
    To better ablate the training efficiency of \datajuicer{} over \datasieve{}, we align \dataGovernor{}s under the same token expenses.
	\item \textbf{Model Scaling}. We scale from CLIP-S to CLIP-L, which has around $10\times$ parameters.
	\item \textbf{Schedule Scaling.} We increase the sampled token from 30 to 80G (50 epochs of 12M data).
\end{itemize}

\begin{figure*}[t]
\centering

\begin{subfigure}{0.5\textwidth}
    \centering
    \includegraphics[width=\textwidth]{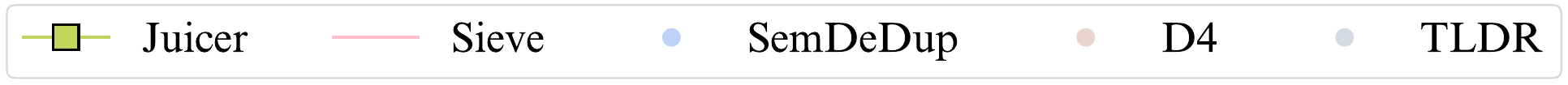}
\end{subfigure}

\begin{subfigure}{0.32\textwidth}
    \centering
    \includegraphics[width=\textwidth]{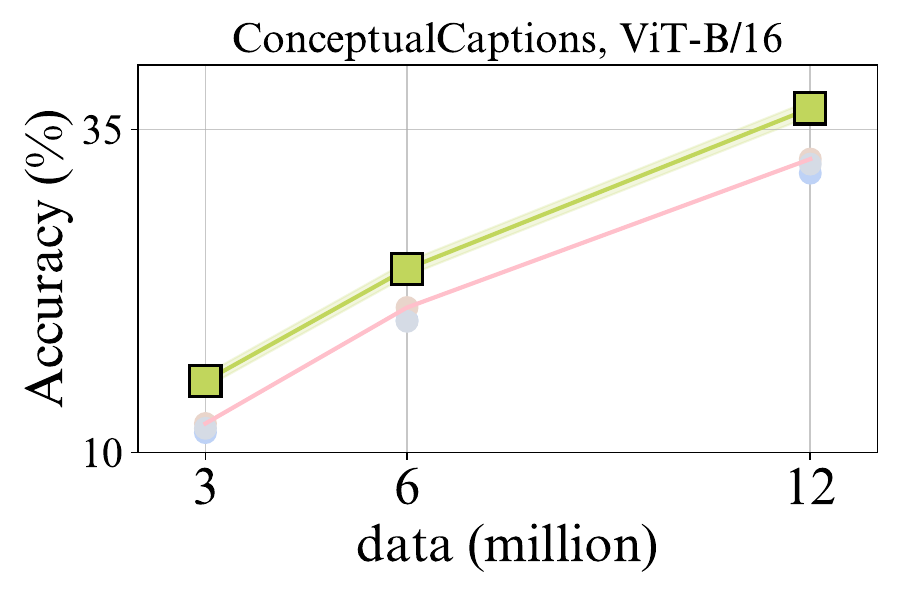}
    \label{fig:datascaling}
\end{subfigure}
%
%
\hfill
\begin{subfigure}{0.32\textwidth}
    \centering
    \includegraphics[width=\textwidth]{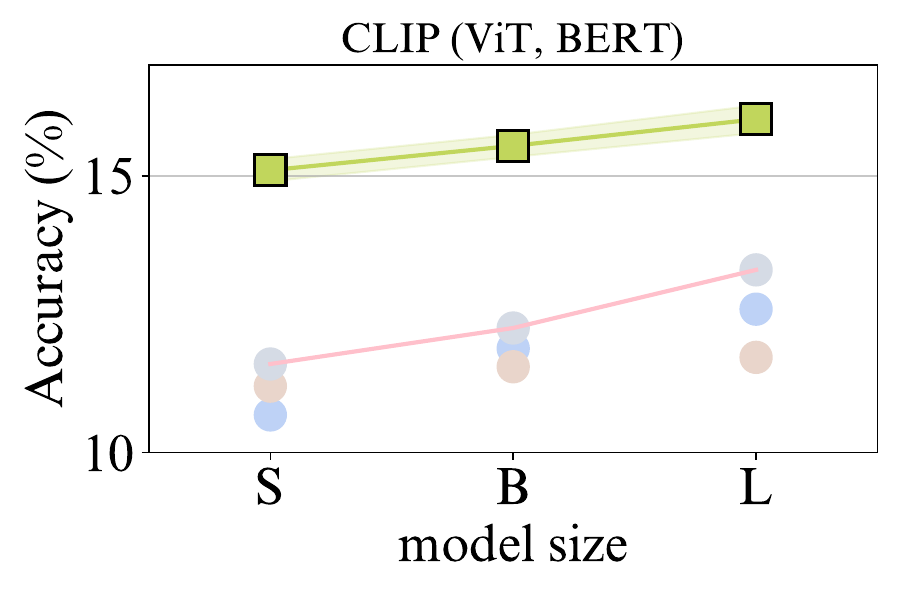}
    \label{fig:modelscaling}
\end{subfigure}
%
%
%
\hfill
\begin{subfigure}{0.32\textwidth}
    \centering
    \includegraphics[width=\textwidth]{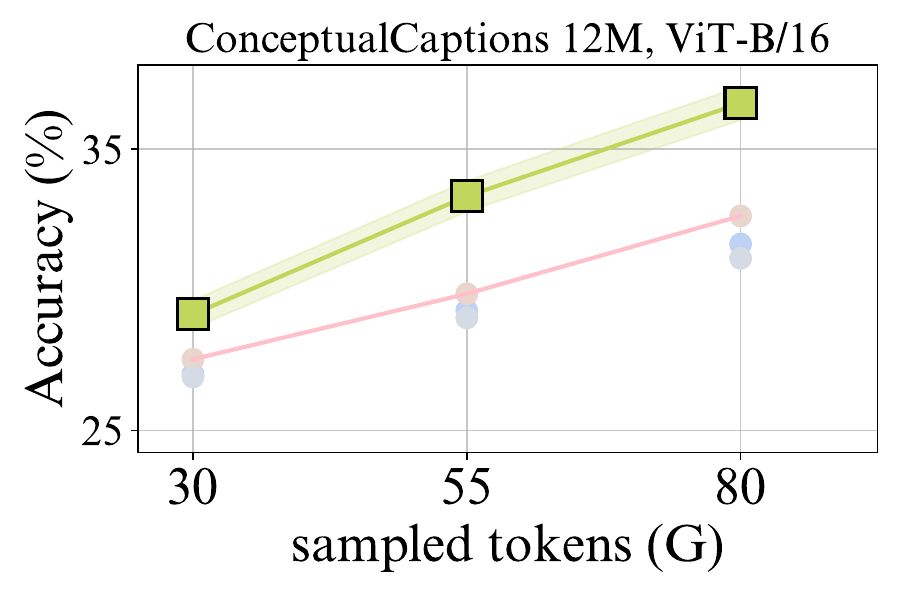}
    \label{fig:schedulescaling}
\end{subfigure}

\vspace{-20pt}
\caption{
    The highest \datasieve{} results are marked in \textcolor{pink}{pink}.
    \textbf{Data Scaling} (Left). 
     With training data scaled from CC3M to CC12M, a clear gap persists throughout training, suggesting \datajuicer{}’s wiser computation investment on informative tokens.
    \textbf{Model Scaling} (Middle). \datajuicer{} outperforms \datasieve{} across CLIP sizes (from S to L), showing large VLP models also learn better on juiced data. 
    \textbf{Schedule Scaling} (Right). We train the same CLIP longer up to 82G sampled tokens (epochs of 12M data).
}

\label{fig:scaling_behavior}

\end{figure*}

\noindent Fig.~\ref{fig:scaling_behavior} studies scaling along one of these three axes at each time while keeping others unchanged. 


\textit{Data Scaling.}
In Fig.~\ref{fig:scaling_behavior} Left, we scale up the training size from 3M to 12M on CLIP(ViT-Base, BERT-Base),
and compare performances of models trained on data generated by \datajuicer{} and \datasieve{} respectively.
The x-axis is the quantity of sampled tokens during training, and the y-axis is the zero-shot accuracy on ImageNet-1K.
Note that with equivalent token sampling volumes,
\datajuicer{} costs less computation while achieving better performance.
A clear gap persists throughout training,
suggesting \datajuicer{}'s wiser computation investment on informative tokens.
With higher per-token marginal returns, \datajuicer{} makes a more suitable \dataGovernor{} than \datasieve{}s under data scaling.

\textit{Model Scaling.}
Fig.~\ref{fig:scaling_behavior} Middle presents \dataGovernor{}s' model scaling behavior,
showing that \datajuicer{}s perform superior than \datasieve{} on larger models.
While the increase in model size enhances performance,
model scaling exacerbates inherent limitations in training data, thus imposing heightened demands in \dataGovernance{}.
Modern MLLMs favor larger image encoders to push through the visual capability bottleneck.
Therefore, the superior performance of \datajuicer{} over \datasieve{} underscores its advantage in supporting advanced foundation models.
The model configurations of ViT-S/B/L are detailed in Appendix~\ref{sec:experiment_details}.

\textit{Schedule Scaling.} Fig.~\ref{fig:scaling_behavior} Right trains longer given the same curated dataset $\mathbb{V}$ and model $\Theta$.
The gain brought by \datajuicer{} is amplified as the sampled data increases.

\subsection{Generalization Evaluation}


\vspace{0.1cm}
\noindent\textbf{Cross-data Evaluation.}
To validate the generalization efficacy for \dataGovernor{}s, we evaluate across various datasets for VLP.
Unprocessed datasets $\mathbb{D}$ are classified into two categories,
clean data filtered with human intervention and web-sourced wild data without cleaning.
Well-cleaned, high-quality examples of the former include CC3M and CC12M, while subsets
YFCC15M (from YFCC100M) and LAION40M (from LAION400M)
are wild data exemplars.

\begin{figure}[htpb]
\centering

\begin{subfigure}{0.32\textwidth}
    \centering
    \includegraphics[width=\textwidth]{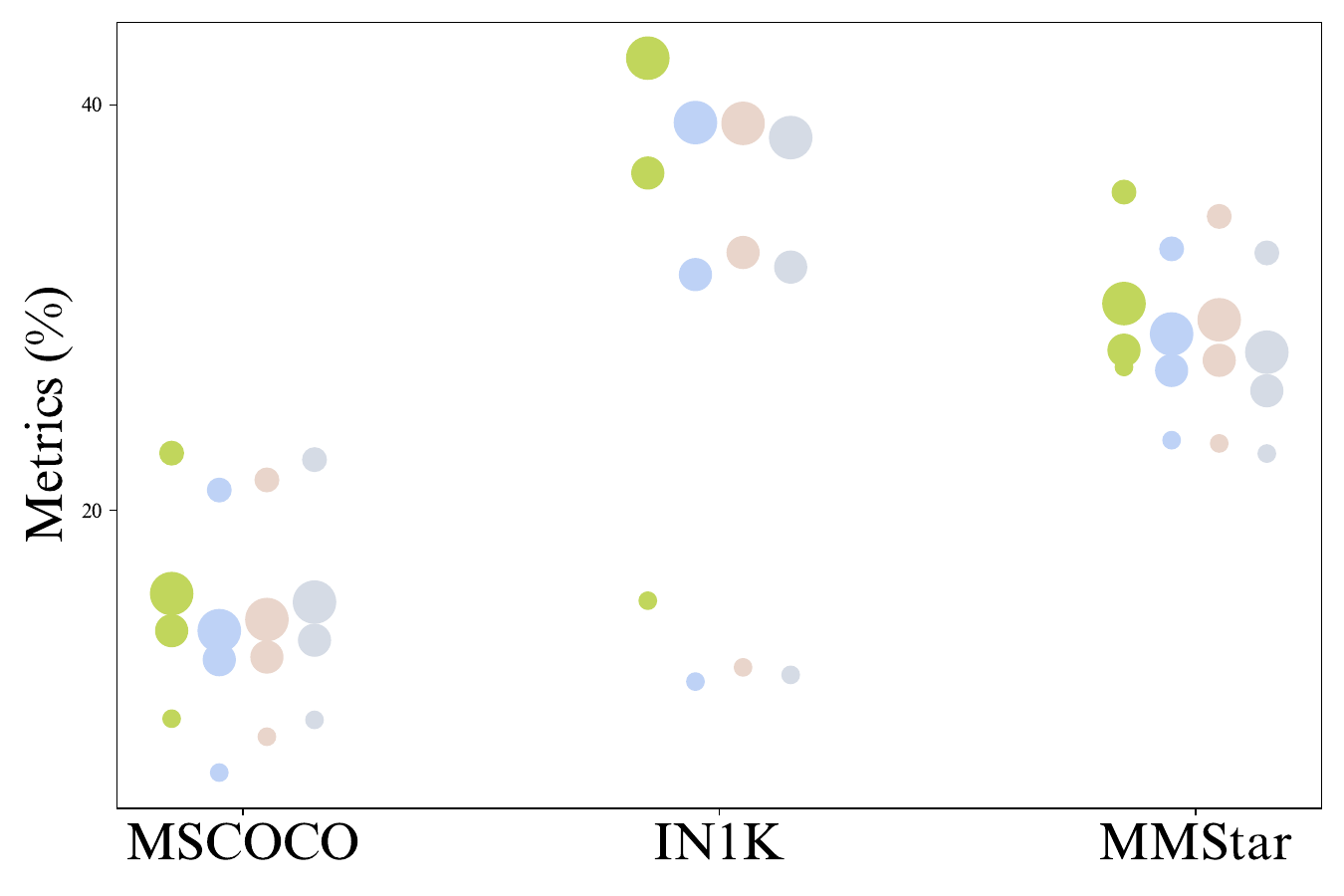}
    \label{fig:first}
\end{subfigure}
\vspace{-15pt}
\begin{subfigure}{0.11\textwidth}
    \centering
    \includegraphics[width=\textwidth]{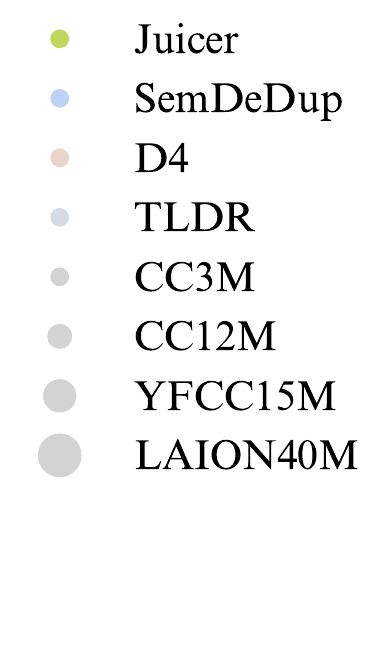}
    \label{fig:first}
\end{subfigure}

\caption{\textbf{Generalization Across Data.} \datajuicer{}
performs better on both well-cleaned datasets and larger noisy datasets (larger markers).
We choose one typical benchmark from each downstream task, {\em i.e.}, MSCOCO for retrieval, ImageNet-1K for classification and MMStar for MLLM evaluation.
}

\label{fig:gen_across_data}
\end{figure}

We use CLIP as the default architecture and evaluate on MSCOCO, ImageNet-1K and MMStar, three representative benchmarks in retrieval, classification and MLLM tasks.
As observed in Fig.~\ref{fig:gen_across_data},
\datajuicer{} outperforms \datasieve{}s across various training data.
The gains on YFCC15M and LAION40M prove the \datajuicer{}'s effectiveness on noisy data sourced from the web.

%
%

\begin{figure}[t]
\centering


\begin{minipage}[c]{0.45\textwidth}
\begin{subfigure}{\textwidth}
    \centering
    \includegraphics[width=\textwidth]{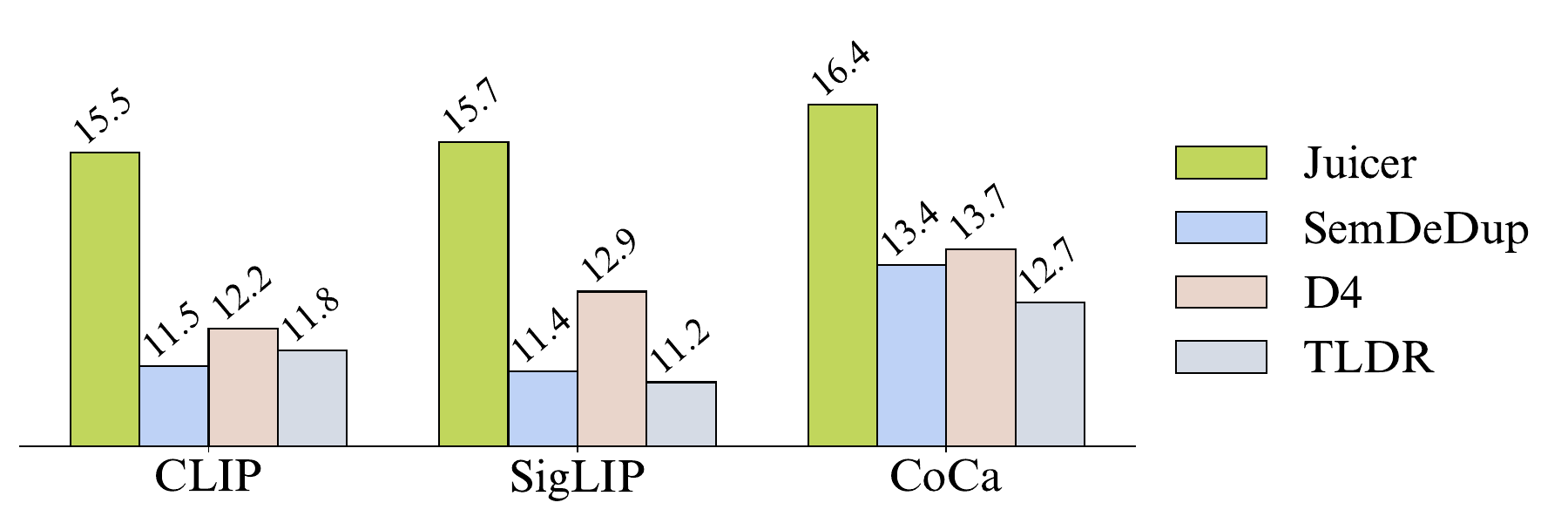}
\end{subfigure}
\end{minipage}
\vspace{-7pt}
\caption{\textbf{Generalization Across Architectures.} For fair comparison, we train models with tokens of equal size, and then report zero-shot performances on ImageNet-1K.
}

\label{fig:gen_across_arch}

\end{figure}

\noindent\textbf{Cross-architecture Evaluation. }
For vision-language pre-training, there are various models that are broadly adopted~\cite{yao2021filip}. These models splits images into patches and embeds these into token sequences, enabling them to leverage the finer-grained \dataGovernor{}. 
As in Fig~\ref{fig:gen_across_arch}, we apply \datajuicer{} to 3 famous model architectures (CLIP~\cite{radford2021learning}, SigLIP~\cite{zhai2023sigmoid}, CoCa~\cite{yu2022coca}),
where \datajuicer{} achieves superior performance than \datasieve{} across all of them.
Therefore we justify \datajuicer{}'s cross-architecture generalization.


\subsection{Ablation Studies}

We perform extensive ablations to dissect designs of \datajuicer{}.
If not stated, experiments use CC3M as the original dataset and conduct a zero-shot evaluation on ImageNet-1K.

\begin{table*}[t]
\centering
\subfloat[
    \textbf{Patch Contribution.} We choose DINO-S to balance performance and computing overhead during \dataGovernance.
    \label{tab:ablation_token_extractor}
]{
    \centering
    \begin{minipage}{0.29\linewidth}{
        \begin{center}
            \tablestyle{4pt}{1.05}
            \begin{tabular}{y{38}x{24}x{24}}
                Model & Top1 & Im/s \\
                \shline
                CLIP & \bf{16.72} & 158.2 \\
                DINO-S & \default{15.54} & \default{\bf 393.4} \\
                DINO-B & 15.58 & 259.3 \\
                DINOv2-L & 15.86 & 190.2 \\
            \end{tabular}
    \end{center}}
    \end{minipage}
}
\hspace{1.5em}
\subfloat[
    \textbf{Patch Selection.} In fact, stratified sampling achieves comparable performances, while we use topk as default for simplicity.
    \label{tab:ablation_patch_selection}
]{
    \centering
    \begin{minipage}{0.29\linewidth}{
        \begin{center}
            \tablestyle{4pt}{1.05}
            \begin{tabular}{y{36}x{24}x{24}}
                Method & Top1 & Top5 \\
                \shline
                uniform & 13.47 & 28.26 \\
                topk & \default{15.54} & \default{{\bf 31.74}} \\
                stratified & \bf{15.61} & 31.28 \\
                mix & 15.47 & 31.17 \\
            \end{tabular}
    \end{center}}
    \end{minipage}
}
\hspace{1.5em}
\subfloat[
    \textbf{Patch Size.} Smaller patches enable more precise patch selection and finer-grained \dataGovernance{}.
    \label{tab:ablation_resolution}
]{
    \centering
    \begin{minipage}{0.29\linewidth}{
        \begin{center}
            \tablestyle{4pt}{1.05}
            \begin{tabular}{y{32}x{26}x{26}}
                Size & Acc & Im/s \\
                \shline
                32 & 15.47 & \bf{407.5} \\
                16 & \default{15.54} & \default{393.4} \\
                14 & \bf{15.94} & 335.6 \\
                 & & \\
            \end{tabular}
    \end{center}}
    \end{minipage}
}
\\
\centering
\subfloat[
    \textbf{Confidence Thresh} should maintain sufficient visual evidence while minimizing noise.
    \label{tab:ablation_confidence_thresh}
]{
    \centering
    \begin{minipage}{0.29\linewidth}{
        \begin{center}
            \tablestyle{4pt}{1.05}
            \begin{tabular}{y{36}x{24}x{24}}
                Thresh ($\epsilon$) & Top1 & Top5 \\
                \shline
                50\% & 15.03 & 31.37 \\
                70\% & \default{\textbf{15.54}} & \default{\textbf{31.74}} \\
                90\% & 14.73 & 30.54 \\
                & & \\
            \end{tabular}
    \end{center}}
    \end{minipage}
}
\hspace{1.5em}
\subfloat[
    \textbf{Caption Enhancement.} Incorporating visual evidence (see Eq.~\ref{equation: class_pred}) improves accuracy.
    \label{tab:ablation_caption_enhancement}
]{
    \centering
    \begin{minipage}{0.29\linewidth}{
        \begin{center}
            \tablestyle{4pt}{1.05}
            \begin{tabular}{y{36}x{24}x{24}}
                Method & Top1 & Top5 \\
                \shline
                none & 11.28 & 24.54 \\
                text-only & 12.76 & 26.12 \\
                concat & 15.31 & 31.56 \\
                rewrite & \default{\bf 15.54} & \default{\bf 31.74} \\
            \end{tabular}
    \end{center}}
    \end{minipage}
}
\hspace{1.5em}
\subfloat[
    \textbf{Full-Patch Tuning} mitigates the train/infer gap but incurs $\mathcal{T}\%$ compute budget.
    \label{tab:ablation_full_patch_tuning}
]{
    \centering
    \begin{minipage}{0.29\linewidth}{
        \begin{center}
            \tablestyle{4pt}{1.05}
            \begin{tabular}{y{32}y{26}x{26}}
                Step & $\mathcal{T}$ (\%) & Top1 \\
                \shline
                0 & 0 & \default{15.54} \\
                2\% & 4$\uparrow$ & 16.02 \\
                4\% & 8$\uparrow$ & 16.35 \\
                8\% & 16$\uparrow$ & \bf{16.72} \\
            \end{tabular}
    \end{center}}
    \end{minipage}
}
\vspace{-0.1cm}
\caption{\textbf{\datajuicer{} Ablation Experiments} use CLIP trained on curated dataset $\mathbb{V}$, and use ImageNet-1k for zero-shot evaluation.
We report Top-1/5 accuracy and fp32 model throughput (im/s) on a V100 GPU.
Our default settings are marked in {\setlength{\fboxsep}{2pt}\colorbox{defaultcolor}{{purple}}}.}
\label{tab:ablations}
\vspace{-0.1cm}
\end{table*}

Patch contributions are quantified based on the contributions of their corresponding tokens to the overall semantics encoded by ViT.
Pretrained ViT from DINO-S best balances performance and inference speed (Tab.~\ref{tab:ablation_token_extractor}).
Despite CLIP's competitive performance, we opt for vision-only foundation models (\textit{e.g.} DINO, MAE) to insulate patch contribution estimation from text interference.

Patch selection applies sampling on scored image patches and forwards chosen ones for training. Stratified sampling selects from patch partitions of different scores and performs slightly better than other sampling (Tab.~\ref{tab:ablation_patch_selection}).
We use topk selection as default to keep the selection user-friendly.

The image patch size decides the granularity of \datajuicer{},
with smaller patches enabling more precise selection.
As shown in Tab.~\ref{tab:ablation_resolution}, models trained with images of smaller patches benefit even more from \datajuicer{}.

Confidence threshold $\epsilon$ is set to extract high-certainty visual evidence from images.
As in Tab.~\ref{tab:ablation_confidence_thresh},
we set the confidence thresh $\epsilon = 0.7$ to avoid compromising caption accuracy.
Users can choose a moderately relaxed threshold to merge more visual semantics to enhance captions.

Caption enhancement incorporates visual semantics from \datajuicer{}'s vision branch to refine and enrich captions with LLM, thereby improving image-caption alignment.
\datajuicer{} degenerates to vision-only \dataGovernor{} when none caption enhancement applied.
Comparison against text-only enhancement and direct object name concatenation demonstrates the benefits of LLM rewrite (Tab.~\ref{tab:ablation_caption_enhancement}).

Full-patch tuning involves all image patches during the late training phase.
Tab.~\ref{tab:ablation_full_patch_tuning} shows that full-patch tuning on last epoch alone can benefit performance on fine-grained visual recognition ({\em e.g.} POPE), costing only 2\% extra computational overhead.
While it is excluded by default for a fair comparison with \datasieve{}, this idea remains optional for users,
to incur tasks demanding all image details.

\definecolor{darkred}{HTML}{c00000}
\definecolor{lightred}{HTML}{e69999}
\definecolor{lighterred}{HTML}{f2cccc}

\begin{figure}[htpb]
\centering
\begin{subfigure}{0.47\textwidth}
    \centering
    \includegraphics[width=\textwidth]{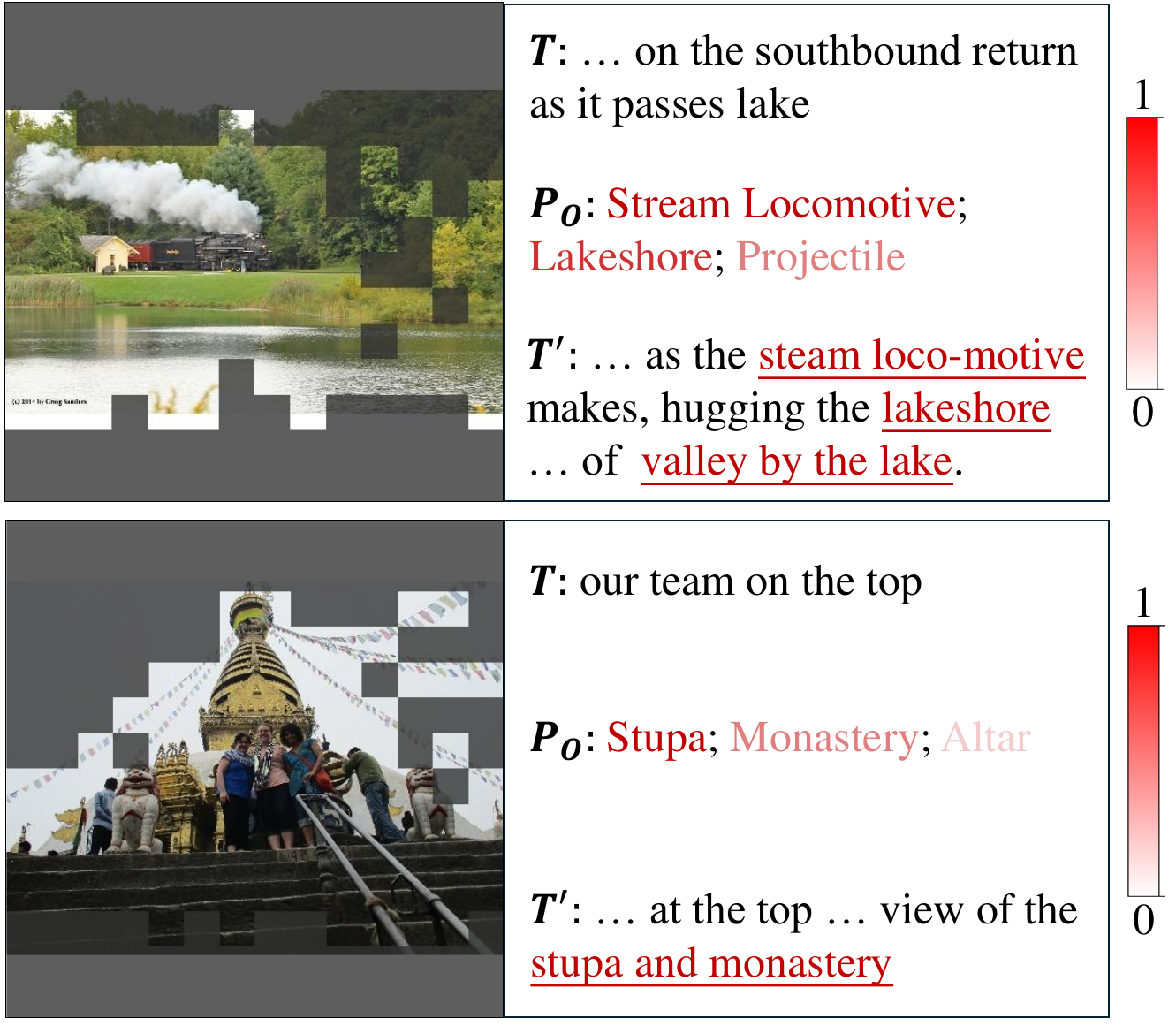}
\end{subfigure}
\vspace{-0.2cm}
\caption{\datajuicer{} reduce visual redundancy by discarding less contributive 50\% image patches as masked in visualization.
We mark visual evidence $P_O$ in red from \textcolor{darkred}{dark} to \textcolor{lightred}{light} based on their confidence. More visualizations are to be found in Appendix ~\ref{sec:more_visualization}.
}
\label{fig:visualization}
\end{figure}


%

\subsection{Visualizations}
Fig.~\ref{fig:visualization} illustrates some modified samples generated by \datajuicer{}, where significant patches are removed through \textbf{patch selection}
and masked with gray regions.
Real-world text descriptions often exhibit imperfections, vague descriptions of objects, omission of key elements, non-descriptive expressions, etc.
We demonstrate \textbf{caption enhancement} as follows:
(1) \textit{Top}: The key elements `Steam Locomotive, Lakeshore' are omitted in the original caption. \datajuicer{} utilizes LLM to infer the scene given the original text and objects predicted from Vision FMs, effectively reconstructing the image's context.
(2) \textit{Bottom}: A significant portion of web images are shared to convey experiences, where the captions are complementary semantics of images
(\textit{e.g.} 'stupa' in the image and 'team' in the caption),
resulting in weak correspondence between the image and the original caption.
\datajuicer{} integrates image semantics with the original caption to form a complete description of the scene.
Accordingly, it is concluded that \datajuicer{} deals with data redundancy and noise effectively.

\vspace{0.2cm}
\section{Conclusion}
The paper presents \datajuicer{} as a novel approach as  \dataGovernance.
By employing a token view, we significantly improve the data quality and model performance. 
With rigorous experiments and massive ablations, our framework shows superior performance in image-text retrieval, classification and a range of dense visual tasks,  surpassing \datasieve{} with substantial improvements.
This paper emphasizes the vital advantage of \datajuicer{} for finer-grained \dataGovernor,
paving the way for future progress in \dataGovernance.

\clearpage
{
    \small
    \bibliographystyle{ieeenat_fullname}
    \bibliography{main}
}

\clearpage
\pagenumbering{Roman}
\setcounter{page}{1}
\maketitlesupplementary
\appendix

\section{Detailed Implementation and Results}
\label{sec:experiment_details}

\newcolumntype{y}[1]{>{\raggedright\arraybackslash}p{#1pt}}
\vspace{5pt}
\noindent\textbf{Pretraining Hyperparamters.}
The standard configuration for our pre-training process is shown bellow.
By default, we employ float32 as the numerical precision.

\begin{table}[htpb]
	\tablestyle{6pt}{1.02}
	\begin{tabular}{y{100}|x{92}}
		Config & Value \\
		\shline
		optimizer & AdamW \\
		base learning rate & 5e-5 \\
		weight decay & 0.05 \\
		optimizer momentum & $\beta_1, \beta_2{=}0.9, 0.999$ \\
		batch size & 1024 \\
		learning rate schedule & cosine decay \\
		warmup epochs & 5 \\
	\end{tabular}
	\label{tab:pretrain_hyperparam}
\end{table}

\noindent\textbf{Model Architecture.}
We detail the CLIP architecture from S to L as follows.

\begin{table}[htpb]
    \tablestyle{6pt}{1.02}
    \begin{tabular}{l|ccccccccc}
        & Learning & Embedding & Input      & \multicolumn{3}{c}{Vision Transformer} & \multicolumn{3}{c}{Text Transformer} \\
        Model & rate & dimension & resolution & layers & width & heads  & layers & width & heads \\
        \shline
        ViT-S/16 & $5 \times 10^{-4}$ & 384 & 224 & 12 & 384 & 6 & 12 & 384 & 6 \\
        ViT-B/16 & $5 \times 10^{-4}$ & 512 & 224 & 12 & 768 & 12 & 12 & 512 & 8 \\
        ViT-L/16 & $5 \times 10^{-4}$ & 768 & 224 & 24 & 1024 & 16 & 12 & 768 & 12 \\
    \end{tabular}
    \label{tab:model_arch}
\end{table}

%

\section{More Visualization}
\label{sec:more_visualization}

\begin{figure*}[htb]
\centering
\includegraphics[width=1\textwidth]{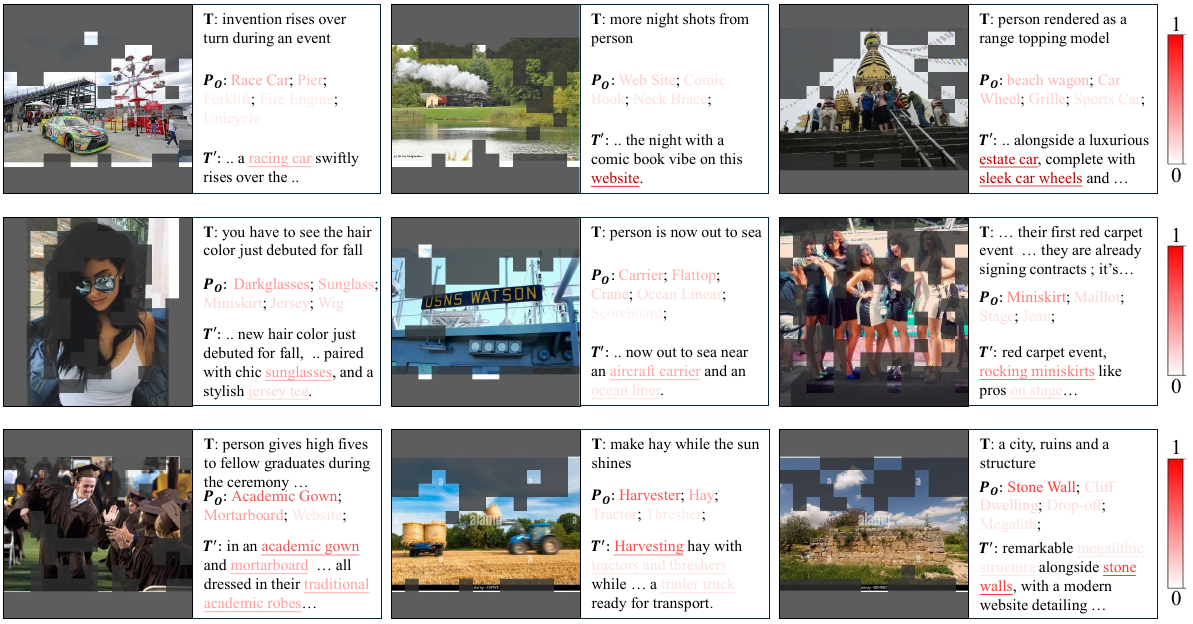}

\caption{\textbf{More Visualizations}.
Image patch elimination reduces the redundant pixels.
As for text enhancement, \datajuicer{} completes the description with detailed class descriptions provided by the vision branch.
The text branch infers the interactions among objects with world knowledge in LLM.
}
\label{fig:appendix_vis}
\end{figure*}

We show the modified samples from \datajuicer{} in Figure~\ref{fig:appendix_vis}.
To demonstrate the caption enhancement for various samples, we group the samples into rows according to the class confidence within images.
Specifically, we categorized samples based on the top confidence scores of class predictions: scores below 30 were denoted as \textcolor{lighterred}{low}, scores between 30 and 60 as \textcolor{lightred}{mid}, and scores above 60 as \textcolor{darkred}{high}.
The confidence scores reflect the model's certainty in identifying objects within the image.

High confidence scores underscore the reliability of
identified objects in images.
We define trustworthy samples as those whose top-confidence objects lie in \textcolor{lightred}{mid} to \textcolor{darkred}{high} confidence ranges.
With more trustworthy classes, the increasing visual evidence coalesces into a cohesive representation of scene semantics.
Take the subfigure(third row, second column) as an example;
the class prediction 'hay' provides objects instead of a comprehensive view.
Combined with 'harvester', we get an explicit scene of tractors and thrashers harvesting hay.
Hence, multiple high-confidence classes make the scene content more specific, enabling the data governor to refine existing captions more confidently.
Mutually verifying visual evidence and their synergistic interactions amplify the utilization of the LLMs's inherent world knowledge, optimizing caption quality.
Rather than modifying sentence structures (e.g., appending), LLMs seamlessly integrate visual evidence into the text descriptions.

When the highest-scoring objects within a sample exhibit \textcolor{lighterred}{low} confidence scores, we prioritize the most salient objects as the primary visual evidence.
And when the texts are non-descriptive, neglecting the primary objects, the visual and textual semantics are likely complementary.
%
%
\datajuicer{}'s text branch $\governorTXT$ typically leverages high-confidence class information to refine captions, such as transforming generic terms (e.g., "car") into more specific descriptors (e.g., "racing car").

To conclude, \datajuicer{} enhances captions with detailed visual evidence from the vision branch and scene comprehension from the text branch.

\vspace{1cm}

\section{Limitation}
\label{sec:limitatoin}
Despite the gains brought by finer-grained \dataGovernance{}, there exist some limitations about \datajuicer{}.
Our token view requires the model to embed data as tokens, \textit{e.g.} image patches, and caption words.
Though compatible with mainstream model architectures nowadays, there are possibilities that it would not benefit novel architectures proposed in the future.

\end{document}